\documentclass{article} % For LaTeX2e
\usepackage{iclr2020_conference,times}

\usepackage{hyperref}
\usepackage{url}

\usepackage{microtype}
\usepackage{graphicx}
\usepackage{subcaption}
\usepackage{booktabs}
\usepackage{wrapfig}
\usepackage{float}
\usepackage{amsmath}
\usepackage{amssymb}
\usepackage{mathtools}
\usepackage{tabu}
\usepackage{import}
\usepackage[]{xcolor}
\usepackage{pifont}% http://ctan.org/pkg/pifont
\usepackage{tikz}
\usetikzlibrary{bayesnet}
\usepackage{sistyle}

\usepackage{cleveref}

\crefformat{section}{\S#2#1#3} % see manual of cleveref, section 8.2.1
\crefformat{subsection}{\S#2#1#3}
\crefformat{subsubsection}{\S#2#1#3}

\newcommand{\cmark}{\color{green} \ding{51}}%
\newcommand{\xmark}{\color{red} \ding{55}}%

\newcommand{\econ}{\textsc{econ}}
\newcommand{\monet}{\textsc{monet}}
\newcommand{\iodine}{\textsc{iodine}}
\newcommand{\genesis}{\textsc{genesis}}
\newcommand{\vae}{\textsc{vae}}
\newcommand{\gan}{\textsc{gan}}
\newcommand{\rbm}{\textsc{rbm}}
\newcommand{\mrbm}{\textsc{m-rbm}}

\newcommand{\x}{\mathbf{x}}
\newcommand{\z}{\mathbf{z}}
\newcommand{\m}{\mathbf{m}}
\newcommand{\s}{\mathbf{s}}
\newcommand{\rr}{\mathbf{r}}
\newcommand{\R}{\mathbb{R}}
\newcommand{\N}{\mathcal{N}}
\newcommand{\LL}{\mathcal{L}}
\newcommand{\E}{\mathbb{E}}
\newcommand{\given}{\: | \:}

\tikzstyle{plate caption} = [caption, node distance=0, inner sep=0pt, below left=-.5em and -1em of #1.south east]

\def\*#1{\mathbf{#1}}
\def\argdot{{\hspace{0.18em}\cdot\hspace{0.18em}}}

\DeclarePairedDelimiterX{\infdivx}[2]{(}{)}{#1\;\delimsize\|\;#2}
\DeclarePairedDelimiterX{\norm}[1]{\lVert}{\rVert}{#1}
\newcommand{\KL}{D_{KL}\infdivx}
\title{Towards causal generative scene models\\ via competition of experts}

\author{Julius von K\"ugelgen\thanks{Equal contribution}\, \thanks{Work done during an internship at Amazon Research T\"ubingen} \,$^{1,2}$, 
 Ivan Ustyuzhaninov\footnotemark[1]\, \footnotemark[2] \,$^{3}$,\\
 \textbf{Peter Gehler\thanks{Joint senior author} \,$^{4}$,
 Matthias Bethge\footnotemark[3] \,$^{3,4}$,
 Bernhard Sch\"olkopf\footnotemark[3] \,$^{1,4}$
 }\\
$^{1}$Max Planck Institute for Intelligent Systems T\"ubingen, Germany\\
$^{2}$Department of Engineering, University of Cambridge, United Kingdom\\
$^{3}$University of T\"ubingen, Germany \\
$^{4}$Amazon T\"ubingen, Germany\\
\texttt{\{jvk,bs\}@tuebingen.mpg.de}, \\
\texttt{\{ivan.ustyuzhaninov,matthias.bethge\}@bethgelab.org}, \\
\texttt{pgehler@amazon.com}\\
}

\iclrfinalcopy % Uncomment for camera-ready version, but NOT for submission.
\begin{document}

\maketitle
\begin{abstract}
Learning how to model complex scenes in a modular way with recombinable components is a pre-requisite for higher-order reasoning and acting in the physical world. However, current generative models lack the ability to capture the inherently compositional and layered nature of visual scenes. While recent work has made progress towards unsupervised learning of object-based scene representations, most models still maintain a global representation space (i.e., objects are not explicitly separated), and cannot generate scenes with novel object arrangement and depth ordering.
Here, we present an alternative approach which uses an inductive bias encouraging modularity by training an ensemble of generative models (\emph{experts}). During training, experts compete for explaining parts of a scene, and thus specialise on different object classes, with objects being identified as parts that re-occur across multiple scenes. Our model allows for controllable sampling of individual objects and recombination of experts in physically plausible ways. In contrast to other methods, depth layering and occlusion are handled correctly, moving this approach closer to a causal generative scene model.
Experiments on simple toy data qualitatively demonstrate the conceptual advantages of the proposed approach.
\end{abstract}

\section{Introduction}
\label{sec:introduction}
Proposed in the early days of computer
vision \cite{grenander1976patternsynthesis,horn1977understandingintensities}, \emph{analysis-by-synthesis} is an approach to the problem of visual scene understanding.
The idea is conceptually elegant and appealing: build a system that is able
to synthesize complex scenes (e.g., by rendering), and then understand
analysis (inference) as the inverse of this process that decomposes new scenes
into their constituent components.
The main challenges in this approach are the need for generative models of objects (and their composition into scenes)
and the need to perform tractable inference given new inputs, including the task to decompose scenes into objects in the first place.
In this work, we aim to learn such as system in an unsupervised way from
observations of scenes alone.

\begin{wrapfigure}{R}{0.5\textwidth}
\centering
\textbf{A}
\hfill
\includegraphics[width=.45\textwidth]{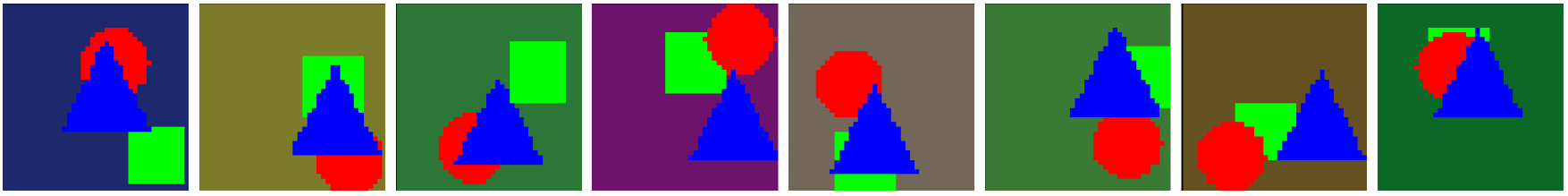}

\vspace{1em}
\textbf{B}
\hfill
\includegraphics[width=.45\textwidth]{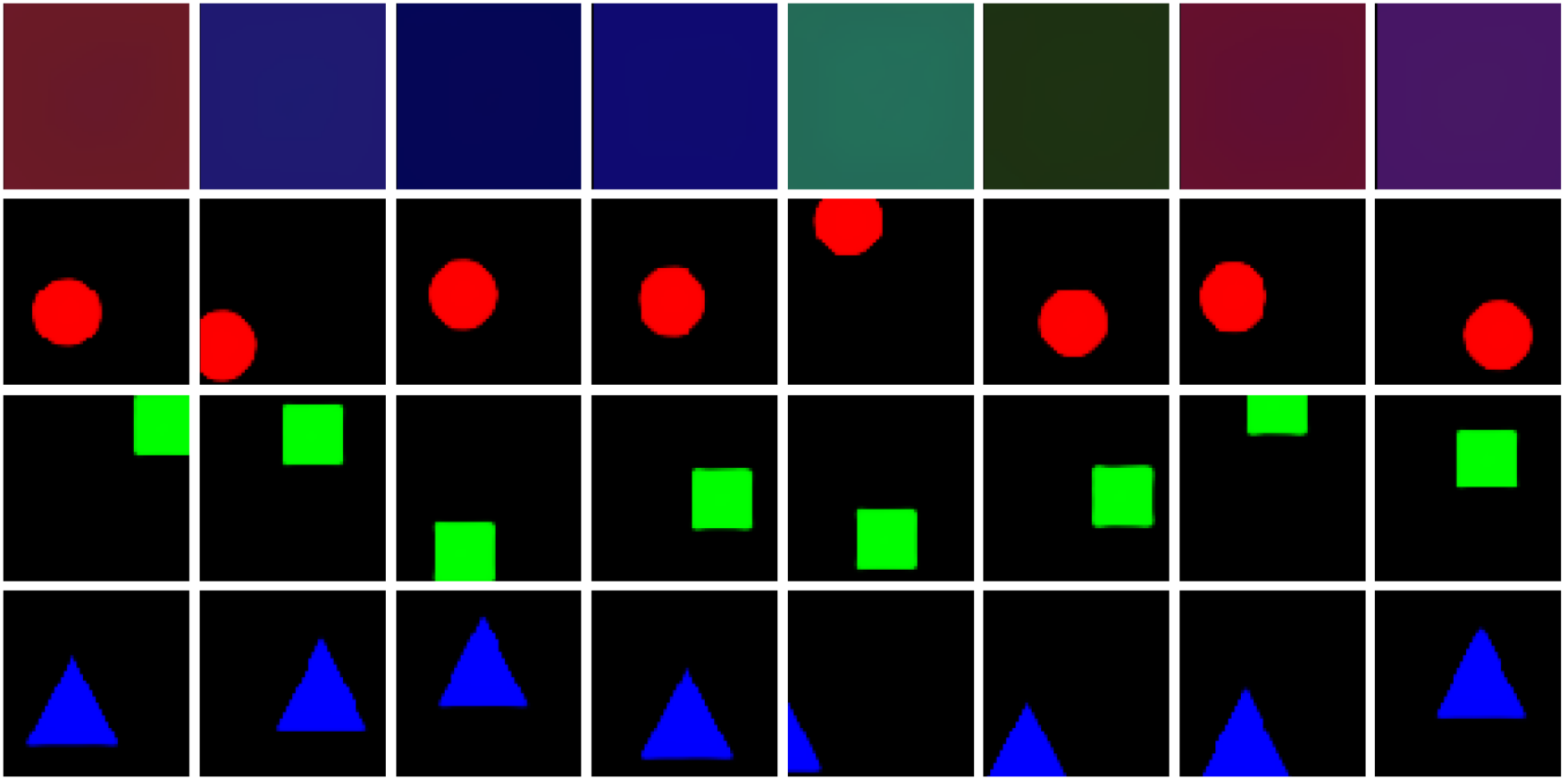}

\vspace{1em}
\hfill
\includegraphics[width=.45\textwidth]{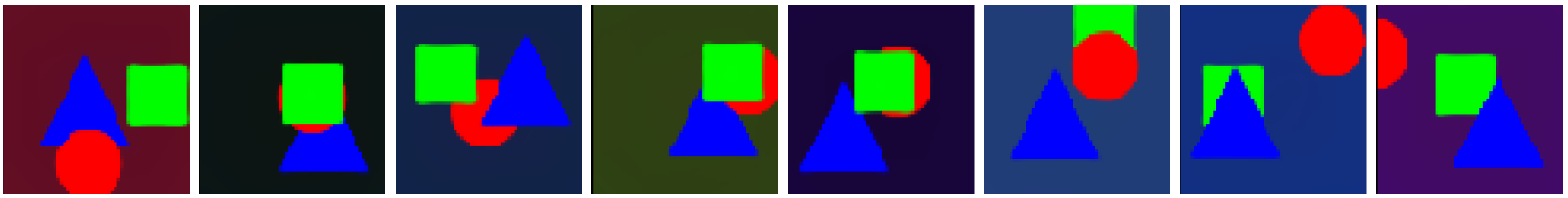}

\hfill
\includegraphics[width = .45\textwidth]{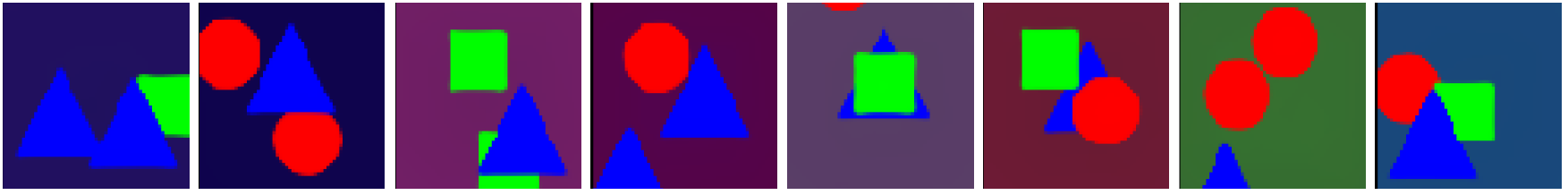}

\textbf{C}
\hfill
\includegraphics[width=.45\textwidth]{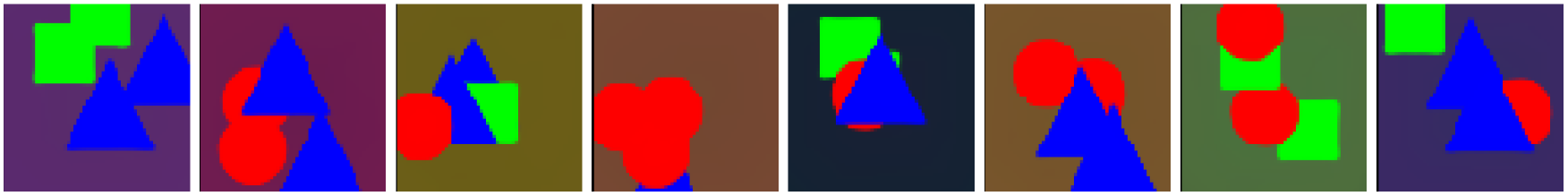}
\caption{Our \econ{} model learns
to decompose training scenes (A) into layers of inpainted
objects. Representing object classes separately allows controllable sampling of
individual objects (B: samples from different experts) which can be recombined
in novel ways (C: compositions sampled by layering the experts in B in the same
order as seen during training (top), or choosing three (middle) or four
(bottom) objects at random).}
\label{fig:samples}
\end{wrapfigure}

While models such as \vae s \citep{kingma2014autoencoding,rezende2014stochastic} and \gan s \citep{GAN} constitute significant progress in generative modelling, these models still lack the ability to capture the compositional nature of reality:
they typically generate entire images or scenes at once, i.e., with a single pass through a large feedforward network.
While this approach works well for objects such as centred faces---and progress has been impressive on those tasks~\cite{karras2019stylegan, karras2019stylegan2}---generating natural scenes containing several objects in non-trivial constellations gets increasingly difficult within this framework due to the combinatorial number of compositions that need to be represented and reasoned about \citep{bau2019seeing}.

Image formation entangles different components in highly non-linear ways, such as \textit{occlusion}. Due to the difficulty of choosing the correct model and the complexity of inference, the task to generate complex scenes containing compositions of objects still lacks success stories.
More training data certainly helps, and progress on generating visually impressive scenes has been substantial \cite{radford2015unsupervised}, but we hypothesize that a satisfactory and robust solution that is not optimized to a relatively well constrained IID (independent and identically distributed) data scenario will require that our models correctly incorporate the (causal) generative nature of natural scenes.

Here, we take some first small steps towards addressing the aforementioned limitations by proposing \econ, a more physically-plausible generative scene model with explicitly compositional structure.
Our approach is based on two main ideas.
The first is to consider scenes as \textit{layered} compositions of (partially) depth-ordered objects.
The second is to represent object classes separately using an ensemble of generative
models, or \emph{experts}.

Our generative scene model consists of a \textit{sequential process} which places independent objects in the scene, operating from the back to the front, so that objects occurring closer to the viewer can occlude those further away.
During inference, this process is reversed:
at each step, experts compete for explaining part of the remaining scene, and only the winning expert is further trained on the explained part \citep{ICM}.
This competition ideally drives each expert to
specialise on representing and generating instances from one, or a few related,
object classes or concepts, and the notion of ``objects'' should automatically emerge as contiguous regions that appear in a stable way across a range of training images.
By decomposing scenes in the reverse order of generation, occluded objects can be inpainted within the already explained regions so that experts can learn to generate full, unoccluded objects which can be recombined in novel ways.

Learning a modular scene representation via object-specific experts has several benefits.
First, each expert only needs to solve the simpler subtask of representing and generating
instances from a single object class---something which current generative
models have been shown to be capable of---while the composition process is treated separately.
Secondly, expert models are useful in their own right as they can be dropped or added, reused and repurposed for other tasks on an individual level.

We highlight the following contributions.
\begin{itemize}
\item We summarise a physically-plausible model of scene generation in \cref{sec:layer_based_model}  and use it to categorise and contrast related scene models and their shortcomings in \cref{sec:related_work}.
\item In \cref{sec:method}, we present \econ, a compositional scene model, which, for a single expert, can be seen as extension of \monet \  \citep{MONET} into a proper generative model (\cref{sec:ELBO}).
\item We introduce modular object representations through separate generators and propose a competition mechanism and objective to drive experts to specialise in \cref{sec:competition_mechanism}.
\item In experiments on synthetic data in \cref{sec:experiments_and_results} we show qualitatively that \econ \ is able to decompose simple scenes into objects, represent these separately, and recombine them in a layer-wise fashion into novel, coherent scenes with arbitrary numbers and depth-orderings of objects.
\item We critically discuss our assumptions and propose extensions for future work in \cref{sec:discussion}.
\end{itemize}

\section{The layer-based model of visual scenes}
\label{sec:layer_based_model}

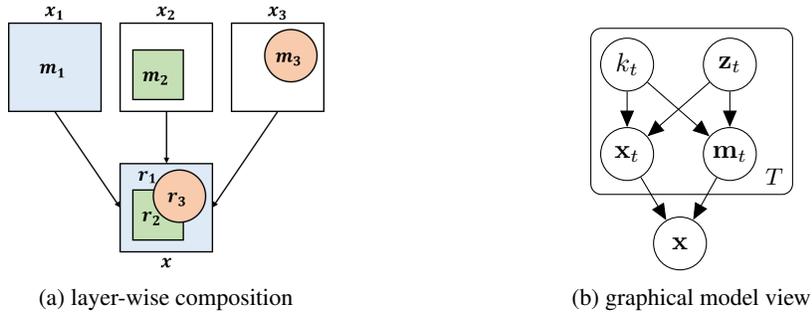
\begin{figure}[]
  \centering
    \begin{subfigure}[b]{0.5\textwidth}
    \centering
    \includegraphics[width=0.6\textwidth, trim={0 8pt 0 0}, clip]{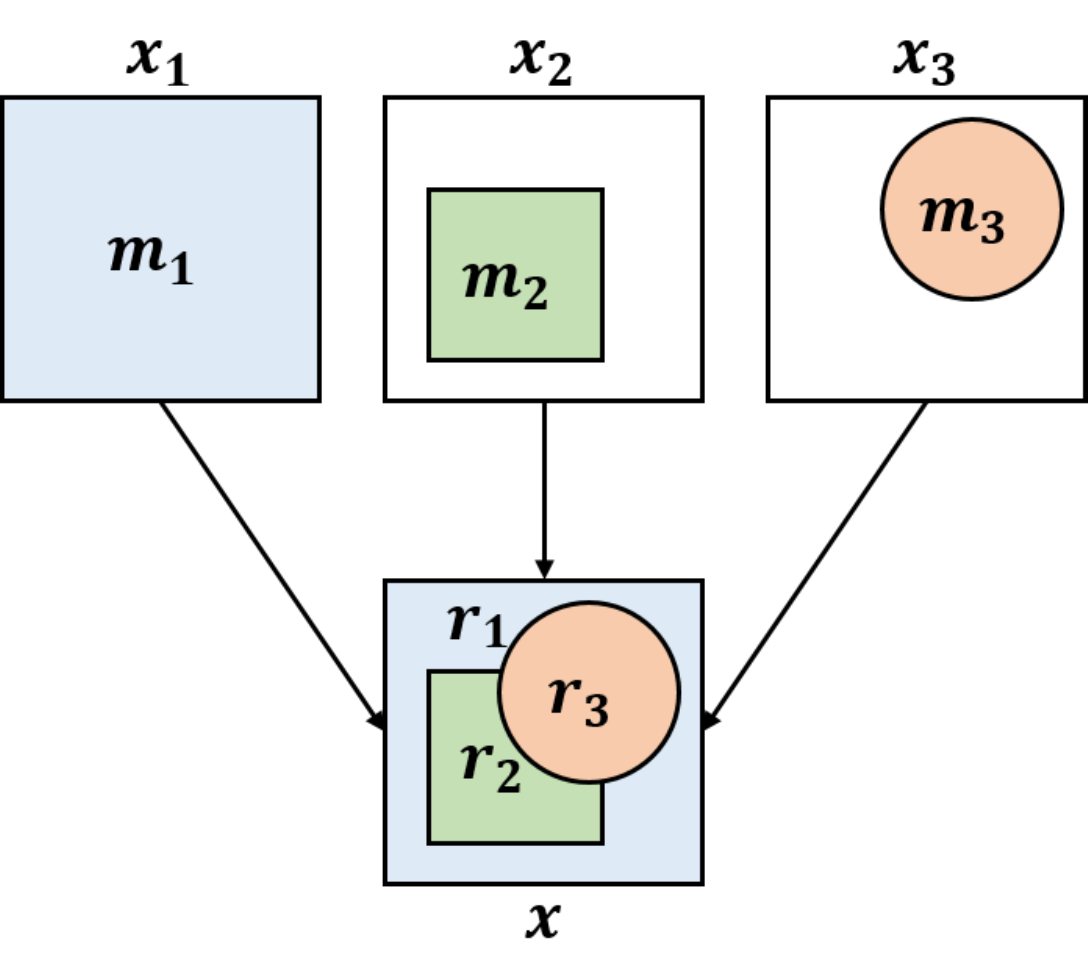}
    \caption{layer-wise composition}
    \label{fig:dead_leaves}
  \end{subfigure}%
  \begin{subfigure}[b]{0.5\textwidth}
    \centering
    \begin{tikzpicture}
      \centering
      \node (k) [latent] {$k_t$};
      \node (z) [latent, right=of k, xshift=-1em] {$\mathbf{z}_t$};
      \node (m) [latent, below=of z, yshift=1.5em] {$\mathbf{m}_t$};
      \node (x_new) [latent, below=of k, yshift=1.5em] {${\mathbf{x}}_t$};
      \node (x) [latent, below=of x_new, xshift=2em, yshift=1.5em] {$\mathbf{x}$};
      \edge {k, z} {m, x_new};
      \edge {m, x_new} {x};
      \plate []{t} {(k) (m) (z) (x_new)} {$T$};
    \end{tikzpicture}
    \caption{graphical model view}
    \label{fig:graphical_model}
  \end{subfigure}
  \caption{Assumed data generating process (dead-leaves model). Independent
  objects $\*x_t$ with shapes $\*m_t$ (drawn from class $k_t$ with properties $z_t$) are placed on the canvas sequentially
  (reflecting depth ordering) and appear in the final composition $\*x$ as
  \textit{dependent}, partially occluded regions $\*r_t$.}
  \label{fig:data_generation}
\end{figure}

To reflect the fact that 2D images are the result of projections of richer 3D
scenes, we assume that data are generated from the well-known \emph{dead leaves
model},\footnote{the name derives from the analogy of leaves falling onto a
canvas, covering whatever is beneath them,} i.e., in a layer-wise fashion, see
Figure \ref{fig:dead_leaves} for an illustration. Starting with an empty canvas $\x =\mathbf{0}$,
an image $\x \in [0,1]^{D \times 3}$ is sequentially generated in $T$ steps.
At each step we sample an object from one of $K$ different classes and place it
on the canvas as follows,
\begin{align*}
  \textbf{for}   \quad &t=1, ..., T: \\
  & k_t \sim p(k_t), \tag*{object class} \\
  & \z_t \sim p(\z_t), \tag*{object properties} \\
  & \m_t \sim p(\m_t \given \z_t, k_t), \tag*{shape} \\
  & \x_t \sim p(\x_t \given \z_t, k_t), \tag*{appearance} \\
  & \x \leftarrow \m_t \odot \x_t + (1 - \m_t) \odot \x, \tag*{place on canvas}
\end{align*}
where $k_t \in \{1,...,K\}$ represents the object class drawn at step $t$;
$\z_t \in \R^L$ is an abstract representation of the object's properties; $\m_t
\in \{0,1\}^D$ is a binary image determining shape; $\x_t \in [0,1]^{D \times
3}$ is a full (unmasked) image containing the object; and $\odot$ denotes
element-wise multiplication. The corresponding graphical model is shown in
Figure \ref{fig:graphical_model}.\footnote{W.l.o.g., we assume that the background corresponds to $\m_1 \odot \x_1$ with
$\m_1 = \mathbf{1}$, see Figure~\ref{fig:dead_leaves}.}

This sequential generation process captures the loss of depth information when
projecting from 3D to 2D and is a natural way of handling occlusion phenomena.
Consequently, sampling from this model is straightforward. We therefore consider
it a more truthful approach to modelling visual scenes than, e.g., spatial
mixture models, in line with~\citet{le2011learning}.

On the other hand, inferring the objects composing a given image $\x$ is
challenging. We will distinguish between \emph{shapes} and \emph{regions} in the
following sense. The unoccluded object shapes $\m_t$, top row in
Figure~\ref{fig:dead_leaves}, remain hidden and only appear in $\x$ via their
corresponding, partially occluded segmentation regions
$\protect{\rr_t \in \{0,1\}^D}$, see the final composition in the bottom row of Figure
\ref{fig:dead_leaves} for an illustration. 
In particular, a region $\rr_t$ is always subset of the corresponding shape pixels $\m_t$.

In addition to the separate treatment of \textit{shapes} $\m_t$
and \textit{regions} $\rr_t$, we also introduce a \textit{scope} variable $\s_t$
to help write the above model in a convenient form. Following \citet{MONET},
$\s_t \in \{0,1\}^D$ is defined recursively as
\begin{equation}
  \s_T := \mathbf{1}, \quad \s_t := \s_{t+1} \odot (1 - \m_{t+1}) \quad \forall t < T.
  \label{eq:scope}
\end{equation}
The scope $\s_t$ at time $t$ contains those parts of the image, which have been
completely generated after $t$ steps and will not be occluded in the subsequent
$T - t$ steps.

With $\s_t$, the regions $\rr_t$ can be compactly defined as
\begin{equation}
  \rr_t = r(\m_t, \ldots, \m_T) := \s_t \odot \m_t, \quad t = 1, \ldots, T.
  \label{eq:segmentation_regions}
\end{equation}
Using these, we can express the final composition as
\begin{equation}
  \label{eq:composition}
  \x = \sum_{t=1}^T \rr_t \odot \x_t.
\end{equation}
While \eqref{eq:composition} may look like a normal spatial mixture model, it is
worth noting the following important point: even though the shapes $\m_t$ are
drawn independently, the resulting segmentation regions $\rr_t$ become
(temporally) dependent due to the layer-wise generation process, i.e., the
visible part of object $t$ depends on all objects subsequently placed on the
canvas. This seems very intuitive and is evident from the fact that the RHS of
\eqref{eq:segmentation_regions} is a function of $\m_{t:T}$.

\section{Related work}
\label{sec:related_work}

\paragraph{Clustering \& spatial mixture models}
One line of work \citep{TAGGER, NEM, RNEM} approaches the \textit{perceptual
grouping} task of decomposing scenes into components by viewing separate regions
$\rr_t$ as clusters. A scene $\x$ is modelled with a spatial mixture model,
parametrised by deep neural networks, in which learning is performed with a
procedure akin to expectation maximisation~\citep[EM;][]{dempster1977maximum}.
The recent \iodine{} model of~\citet{IODINE} instead uses a refinement network
\citep{marino2018iterative} to perform iterative amortised variational inference
over independent scene components which are separately decoded and then combined
via a softmax to form the scene. While \iodine{} is able to \textit{decompose} a
given scene, it cannot \textit{generate} coherent samples of new scenes because
dependencies between regions $\rr_t$ due to layering are not explicitly captured
in its generative model.

This shortcoming of \iodine{} has also been pointed out by~\citet{GENESIS} and
addressed in their \genesis{} model, which explicitly models dependencies
between regions via an autoregressive prior over $\rr_{1:T}$.  While this
\textit{does} enable sampling of coherent scenes which look similar to training
data, \genesis{} still assumes an additive, rather than layered, model
of scene composition. As a consequence, the resulting entangled component samples contain holes and partially occluded
objects and cannot be easily layered and recombined as shown in
Figure~\ref{fig:samples} (e.g., to generate samples with exactly two circles and
one triangle).

\paragraph{Sequential models}
Our work is closely related to sequential or recurrent approaches to image
decomposition and generation
\citep{mnih2014recurrent,gregor2015draw,AIR,SQAIR,yuan2019generative}. In
particular, we build on the recent \monet{} model for scene decomposition of
\citet{MONET}. \monet{} combines a recurrent attention network with a VAE which
encodes and reconstructs the input within the selected attention regions
$\rr_t$ while unconstrained to inpaint occluded parts outside $\rr_t$.

We extend this approach in two main directions. Firstly, we turn \monet{} into a
proper \textit{generative} model\footnote{in its original form, it is a conditional
model which does not admit a canonical way of sampling new scenes} which
respects the layer-wise generation of scenes described in
\cref{sec:layer_based_model}. Secondly, we explicitly model the discrete
variable $k$ (object class) with an \textit{ensemble} of class-specific VAEs (the \textit{experts})---as opposed to within a single large encoder-decoder architecture as in
\iodine{}, \genesis{} or \monet{}. 
Such specialisation allows to control object constellations in new, but scene-consistent ways.

\paragraph{Competition of experts}
To achieve specialisation on different object classes in our model, we build on ideas from previous work using competitive training of experts \citep{jacobs1991competitive}.
More recently, these ideas have been
successfully applied to tasks such as lifelong learning \citep{aljundi2017expert}, learning
independent causal mechanisms \citep{ICM}, training mixtures of generative
models \citep{locatello2018competitive}, as well as to dynamical systems via
sparsely-interacting recurrent independent mechanisms \citep{RIM}.

\paragraph{Probabilistic RBM models}
The work of~\citet{le2011learning} and \citet{heess2012learning} introduced probabilistic
scene models that also reason about occlusion. \citet{le2011learning} combine
restricted Boltzmann machines (\rbm s) to generate masks and shape separately for
every object in the scenes into a masked \rbm{} (\mrbm) model. Two variants are
explored: one that respects a depth ordering and object occlusions,
 derived from similar arguments as we have put forward in the
introduction; and a second model which uses a softmax combination akin to the spatial
mixture models used in \iodine{} and \genesis, although the authors argue it
makes little sense from a modelling perspective. Inference is implemented as
blocked Gibbs sampling with contrastive divergence as a learning objective.
Inference over depth ordering is done exhaustively, that is, considering every
permutation---as opposed to greedily using competition as in this work. Shortcomings of the model are mainly the limited expressiveness of
\rbm s (complexity and extent), as well as the cost of inference. Our work can be
understood as an extension of the \mrbm{} formulation using VAEs in
combination with attention, or segmentation, models.

\begin{table}[]
  \caption{Comparison with related unsupervised scene decomposition and generation models.}
  \label{tab:related_work}
  \begin{center}
    \vskip -0.1in
    \begin{small}
      {\tabulinesep=.25em
      \begin{tabu} to \columnwidth {X[6, m, l] X[1, m, c] X[1, m, c] X[1, m, c] X[1, m, c] X[1, m, c] }
        \toprule
        & \textsc{monet} & \textsc{iodine} & \textsc{genesis}  &\mrbm{} & \econ \\
        \midrule
        \textit{decompose scenes into objects and reconstruct} & \cmark & \cmark & \cmark & \cmark & \cmark\\
        \textit{generate coherent scenes like training data} & \xmark & \xmark & \cmark & \cmark & \cmark \\
        \textit{controllably recombine objects in novel ways} & \xmark & \xmark & \xmark & \cmark & \cmark \\
        \textit{efficient (amortised) inference} & \cmark & \cmark & \cmark  & \xmark & \cmark \\
        \bottomrule
      \end{tabu}
      }
    \end{small}
  \end{center}
\end{table}

\paragraph{Vision as inverse graphics \& probabilistic programs}
Another way to programmatically introduce information about scene composition is
through analysis-by-synthesis, see~\citet{bever2010analysisbysynthesis} for an
overview. In this approach, the synthesis (i.e., generative) model is fully
specified, e.g., through a graphics renderer, and inference becomes the inverse
task, which poses a challenging optimisation problem. Probabilistic programming
is often advocated as a means to automatically compile this inference task; for
instance, \textsc{picture} has been proposed by~\citet{kulkarni2015picture}, and
combinations with deep learning have been explored by \citet{wu2017neural}. This
approach is sometimes also understood as an instance of Approximate Bayesian
Computation~\citep[ABC;][]{dempster1977maximum} or likelihood-free inference.
While conceptually appealing, these methods require a detailed specification of
the scene generation process---something that we aim to learn in an unsupervised
way. Furthermore, gains achieved by a more accurate scene generation process are
generally paid for by complicated inference, and most methods thus rely on
variations of MCMC sampling
schemes~\citep{jampani2015informedsampler,wu2017neural}.

\paragraph{Supervised approaches} There is a body of work on augmenting
generative models with ground-truth segmentation and other supervisory
information. \citet{turkoglu2019layer} proposed a layer based model to add
objects onto a background, \citet{ashual2019specifying} proposed a
scene-generation method allowing for fine grained user control,
\citet{karras2019stylegan, karras2019stylegan2} have achieved impressive image
generation results by exclusively training on a single class of objects. The key
difference of these approaches to our work is that we exclusively focus on
\emph{unsupervised} approaches.

\section{Ensemble of competing object nets (\econ)}
\label{sec:method}

We now introduce \econ{} (for \textbf{E}nsemble of \textbf{C}ompeting
\textbf{O}bject \textbf{N}etworks), a causal generative scene model which
explicitly captures the compositional nature of visual scenes. On a high level,
the proposed architecture is an ensemble of generative models, or
\emph{experts}, designed after the layer-based scene model described in
\cref{sec:layer_based_model}. During training, experts compete to sequentially
explain a given scene via attention over image regions, thereby specialising on
different object classes. We perform variational
inference~\citep{jordan1999introduction}, amortised within the popular VAE
framework~\citep{kingma2014autoencoding, rezende2014stochastic}, and use
competition to greedily maximise a lower bound to the conditional likelihood
w.r.t. object identity.

\subsection{Generative model}
We adopt the generative model $p$ described in \cref{sec:layer_based_model},
parametrise it by $\theta$, and assume that it factorises over the graphical
model in Figure~\ref{fig:graphical_model} (i.e., assuming that objects at
different time steps are drawn independently of each other). We model $p(k_t)$
with a categorical distribution,\footnote{though we will generally condition on
$k_t$, see \cref{sec:objective} for details,} and place a unit-variance
isotropic Gaussian prior over $\z_t$,
\begin{equation*}
  p(\z_t) = p(\z)= \N(\*0, I) \quad t = 1, \ldots, T.
\end{equation*}

Next, we parametrise $p(\m_t \given k, \z)$ and $p(\x_t \given k, \z)$ using $K$
\emph{decoders}
$f_1, \ldots, f_K: \mathbb{R}^L \rightarrow [0,1]^{D \times 3} \times [0,1]^D$
with respective parameters $\theta_1, \ldots, \theta_K$.\footnote{$K$ is a
hyperparameter ($K$-- 1 object classes and background) which
has to be chosen domain dependently.} These compute object means and mask probabilities
$\big(\mu_{\theta_k}(\*z), \tilde{\*m}_{\theta_k}(\*z)\big)=f_k(\*z; \theta_k)$
which determine pixel-wise distributions over $\*m_t$ and $\*x_t$ via
\begin{align}
  \label{eq:generative_mask_dist}
  p_\theta(\m_t \given \z, k) & = \text{Bernoulli} \left( \tilde{\m}_{\theta_k}(\z) \right), \\
  \label{eq:generative_component_dist}
  p_\theta(\x_t \given \z, k) & = \N \big( \x_t \given \mu_{\theta_k}(\z), \sigma_x^2 I \big),
\end{align}
where $t = 1, \ldots, T$ and $\sigma_x^2$ is a constant variance.

We note at this point that, while other handlings of the discrete variable $k$
are possible, we deliberately opt for $K$ separate decoders: (i) as an inductive
bias encouraging modularity; and (ii) to be able to controllably sample
individual objects and recombine them in novel ways.

Finally, we need to specify a distribution over $\x$. Due to its layer-wise
generation, this is tricky and most easily done in terms of the visible regions
$\rr_t$. From \eqref{eq:composition}, \eqref{eq:generative_component_dist}, and
linearity of Gaussians it follows that, pixel-wise,
\begin{equation}
  p_\theta(\x \given \rr_{1:T}, \z_{1:T}, k_{1:T}) = \N \Big( \x \: \Big| \: \sum_{t=1}^T \rr_t \odot \mu_{\theta_{k_t}}(\z_t), \sigma_x^2I \Big).
  \label{eq:generative_scene_dist}
\end{equation}

Similarly, one can show from \eqref{eq:scope}, \eqref{eq:segmentation_regions}, and \eqref{eq:generative_mask_dist} that  $\*r_t$ depends on $\*r_{(t+1):T}$ only via $\*s_t$, and that
\begin{equation}
  p_\theta(\rr_t \given \s_t, \z, k) = \text{Bernoulli} \left( \s_t \odot \tilde{\m}_{\theta_{k}}(\z) \right),
  \label{eq:generative_region_dist}
\end{equation}
for $t = 1, \ldots, T$; see Appendix \ref{app:derivations} for detailed derivations.

The class-conditional joint distribution then factorises as,
\begin{equation}
    p_\theta(\x, \rr_{1:T}, \z_{1:T}|k_{1:T}) =
    p_\theta(\x \given \rr_{1:T}, \z_{1:T}, k_{1:T}) \prod_{t=1}^T p_\theta(\rr_{t} \given \s_{t}, \z_{t}, k_{t}) p(\z_{t}).
\end{equation}
Conditioning on $k_{1:T}$ is motivated by our inference procedure, see
\cref{sec:objective}. Moreover, we express $p$ in terms of  the segmentation
regions $\rr_t$ as only these are visible in the final composition which makes
is easier to specify a distribution over $\x$. Note, however, that while we will
perform \textit{inference over regions} $\rr_{1:T}$, we will learn to
\textit{generate full shapes} $\m_{1:T}$ which are consistent with the inferred
$\rr_{1:T}$ when composed layer-wise as captured in
\eqref{eq:generative_region_dist}, thus respecting the physical
data-generating process.

\begin{figure}
\centering
  \includegraphics[width=0.75\textwidth]{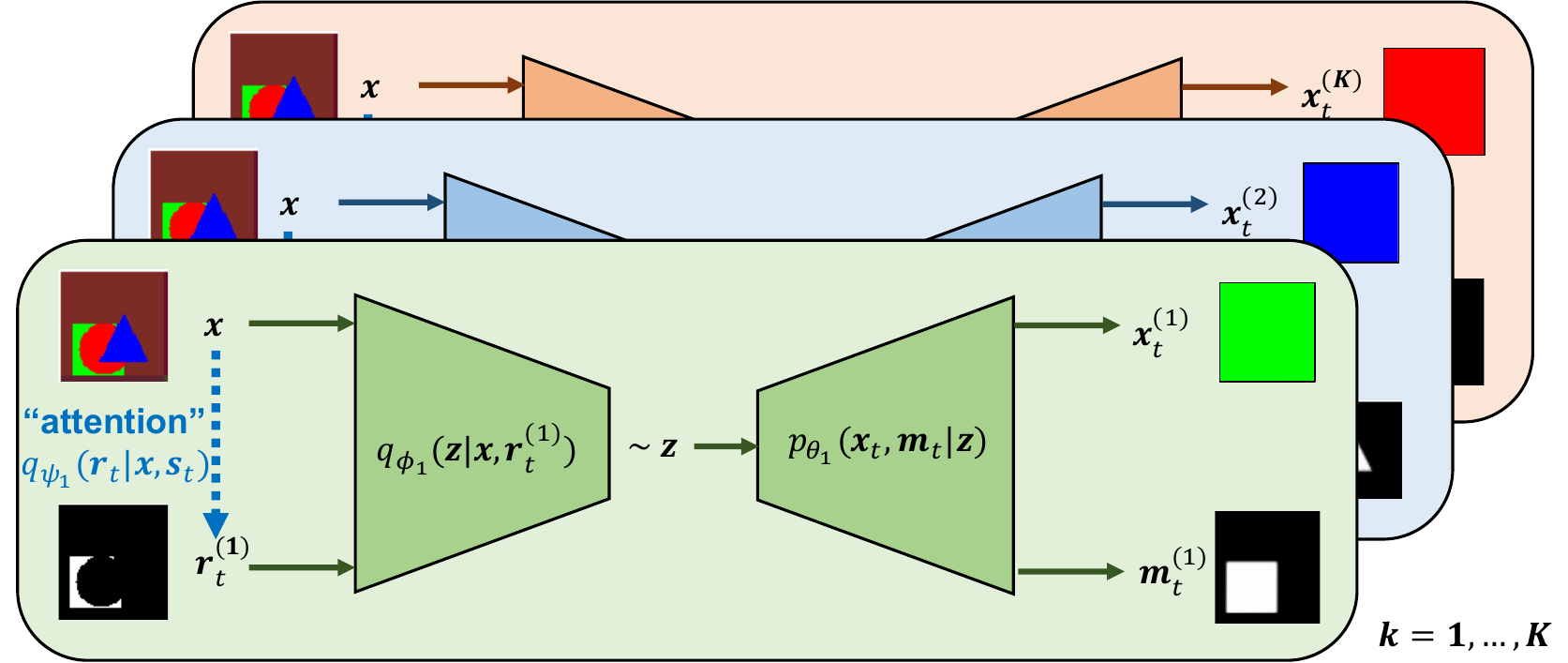}
  \caption{\econ{} architecture: ensemble of $K$ competing experts.
    Each expert consists of (i) an attention network which selects image regions
    $\rr_t$; (ii) an encoder which maps the image within the attended region
    to a latent code $\z$; and (iii) a decoder which reconstructs both an
    object $\x_t$ and its unoccluded shape $\m_t$.
    A competition mechanism determines the winning expert at each step.}
\label{fig:architecture}
\end{figure}

\subsection{Approximate posterior}
\label{sec:approximate_posterior}
Since exact inference is intractable in our model, we approximate the
posterior over $\z_{1:T}$ and $\rr_{1:T}$  with the following variational distribution $q$
parametrised by $\phi$ and $\psi$,
\begin{equation*}
  q_{\phi,\psi} (\rr_{1:T}, \z_{1:T} \given \x, k_{1:T}) =
    \prod_{t=1}^T
      q_{\psi}(\rr_t \given \x, \s_t, k_t)
      q_{\phi}(\z_t \given \x, \rr_t, k_t).
\end{equation*}
As for the generative distribution, we model dependence on $k_t$ using $K$
modules with separate parameters $\{\phi_1, \psi_1\}$, ..., $\{\phi_K,
\psi_K\}$. These inference modules consist of two parts.

\textit{Attention nets} $a_1, ..., a_K:  [0,1]^{D \times 3} \times [0,1]^D
\rightarrow [0,1]^D$ compute region probabilities $\tilde{\rr}_{\psi_k}(\x,\s) =
a_{k}(\x, \s; \psi_k)$ and amortise inference over regions $\rr_t$ via
\begin{equation}
  q_{\psi}(\rr_{t} \given \x, \s_t, k_t) =
    \text{Bernoulli} \left( \tilde{\rr}_{\psi_k}(\x, \s_t) \right)\quad \forall t.
  \label{eq:attention_net}
\end{equation}

\textit{Encoders} $g_1, ..., g_K: [0,1]^{D \times 3} \times [0,1]^D \rightarrow
\R^{L \times 2}$ compute means and log-variances $\left( \mu_{\phi_k}(\x, \rr_t),
\log \sigma^2_{\phi_k}(\x, \rr_t) \right) = g_k(\x, \rr_t; \phi_k)$ which
parametrise distributions over $\z_t$ via
\begin{equation*}
  q_{\phi}(\z_{t} \given \x, \rr_t, k) =
    \N \left(
      \z_t \given \mu_{\phi_k}(\x, \rr_t),
      \sigma^2_{\phi_k}(\x, \rr_t) I
    \right) \quad \forall t.
  \label{eq:encoders}
\end{equation*}

We refer to the collection of $f_k(\argdot;\theta_k)$, $a_k(\argdot;\psi_k)$,
and $g_k(\argdot;\phi_k)$ for a given $k$  as an \textit{expert} as it implements
all computations (generation and inference) for a specific object class---see
Figure \ref{fig:architecture} for an illustration.

\section{Inference}
\label{sec:objective}

Due to the assumed sequential generative process, the natural order of inference
is the reverse ($t=T, \ldots, 1$), i.e., foreground objects should be explained
first and the background last. This is also captured by the dependence of
$\rr_t$ on $\rr_{(t+1):T}$ via the scope $\s_t$ in $q_\psi$.

Such entanglement of scene components across composition steps makes inference
over the entire scene intractable. We therefore choose the following greedy
approach. At each inference step $t=T, \ldots, 1$, we consider explanations from
all possible object-classes $(k_t=1, \ldots, K)$---as provided by our ensemble
of experts via attending, encoding and reconstructing different parts of the
current scene---and then choose the best fitting one. This offers an intuitive
foreground to background decomposition of an image as foreground objects should
be easier to reconstruct.

Concretely, we first lower bound the marginal likelihood conditioned on
$k_{1:T}$, $p_\theta(\x \given k_{1:T})$, and then use a competition mechanism
between experts to determine the best $k$. We now describe this inference
procedure in more detail.

\subsection{Objective: class-conditional ELBO}
\label{sec:ELBO}
First, we lower bound the class-conditional
model evidence $p_\theta(\x \given k_{1:T})$ using the approximate posterior $q$
as follows (see Appendix \ref{app:derivations} for a detailed derivation):
\begin{align*}
&\log p_\theta(\x \given k_{1:T})
\geq
\LL(\theta, \psi, \phi \given k_{1:T})
:=
-\sum_{t=1}^T
   \E_{q_{\psi}(\s_{t} \given \x, k_{(t+1):T})}
   \left( \LL_{\x,t} + \LL_{\z,t}+ \LL_{\rr,t} \right), \\
& \LL_{\x, t}
  :=
  \E_{q_{\psi}(\rr_{t} | \x, \s_t, k_{t}) q_{\phi}(\z_{t} \given \x, \rr_{t}, k_{t})}
 \left[
 \frac{\rr_t}{2\sigma_{\x}^2} \odot
\left(
\x - \mu_{\theta_{k_t}}(\z_t)
\right)^2
 \right]
\\
 &\mathcal{L}_{\*z, t}
:=
\mathbb{E}_{q_{\psi}(\*r_{t}|\*x, \*s_t, k_{t})}
\KL[\Big]{q_{\phi}(\*z_{t}|\*x, \*r_{t}, k_{t})}{ p(\*z)}
\\
&\mathcal{L}_{\*r, t}
:=
\mathbb{E}_{q_{\psi}(\*r_{t} |\*x,\*s_t, k_{t}) q_{\phi}(\*z_{t} |\*x, \*r_{t}, k_{t})}
 \bigg[
 \log \frac{q_{\psi}(\*r_{t}|\*x,\*s_t, k_{t})}{p_\theta(\*r_{t}|\*s_t, \*z_{t}, k_{t})}
 \bigg]
\end{align*}

Next, we use the reparametrization trick of \citet{kingma2014autoencoding} to
replace expectations w.r.t. $q_{\phi}(\z_{t} \given \x, \rr_{t}, k_{t})$ by a
Monte Carlo estimate using a single sample drawn as:
\begin{equation*}
  \tilde{\*z}_t =  \mu_{\phi_{k_t}}(\*x, \*r_t) + \sigma_{\phi_{k_t}}(\*x, \*r_t)\odot \epsilon, \quad \epsilon \sim \mathcal{N}(\*0, I).
\end{equation*}

Finally, we approximate expectations w.r.t. $q_{\psi}(\rr_{t} \given \x, \s_t,
k_{t})$ in $\LL_{\x, t}$ and $\LL_{\z, t}$ using the Bernoulli means
$\tilde{\rr}_{\psi_{k_t}}(\x, \s_t)$. We opt for directly using a continuous
approximation and against sampling discrete $r$'s (e.g., using continuous
relaxations to the Bernoulli distribution \citep{maddison2017concrete,
jang2017categorical}) as our generative model does not require the ability to
directly sample regions. (Instead, we sample $\z$'s and decode them into
unoccluded shapes which can be combined layer-wise to form scenes.)

With these approximations, we obtain the estimates
\begin{align}
\label{eq:rec_loss}
\Hat{\LL}_{\x,t}
 & :=
 \frac{\tilde{\rr}_{\psi_{k_t}}(\x, \s_t)}{2\sigma_{\x}^2} \odot
\left(
\x - \mu_{\theta_{k_t}}(\tilde{\z}_t)
\right)^2,\\
\label{eq:latent_KL}
\Hat{\LL}_{\z, t}
& :=
\KL[\Big]{q_{\phi} \left( \z_{t} \given \x, \tilde{\rr}_{\psi_{k_t}}(\x, \s_t), k_{t} \right)}{ p(\z)}, \\
\label{eq:mask_KL}
\hspace{-2cm} \Hat{\LL}_{\rr, t}
& :=
\KL[\Big]{q_{\psi}(\rr_{t} \given \x,\s_t, k_{t})}{p_\theta(\rr_{t} \given \s_t, \tilde{\z}_{t}, k_{t})},
\end{align}
which we combine to form the \textit{learning objective}
\begin{equation}
    \Hat{\LL}(\theta, \psi, \phi \given k_{1:T})= - \sum_{t = 1}^T \left(\Hat{\LL}_{\x,t} +\beta \Hat{\LL}_{\z, t} + \gamma \Hat{\LL}_{\rr, t} \right),
    \label{eq:learning-objective}
\end{equation}
where $\beta, \gamma$ are hyperparameters. Note that for $\beta,
\gamma > 1$, \eqref{eq:learning-objective} still approximates a valid lower bound.

\paragraph{Generative model extension of \monet{} as a special case}
For $K=1$ (i.e., ignoring different object classes $k_t$ for the moment), our
derived objective \eqref{eq:learning-objective} is similar to that used by
\citet{MONET}. However, we note the following crucial difference in
\eqref{eq:mask_KL}: in our model, reconstructed attention regions
$\tilde{\m}_{\theta_{k}}(\z)$ are multiplied by $\s_t$ in the $p_\theta$ term of
the KL, see \eqref{eq:generative_region_dist}. This implies that the generated
shapes $\m_t$ are constrained to match the attention region $\rr_t$ only within
the current scope $\s_t$, so that---unlike in \monet{}---the decoder is not
penalised for generating entire unoccluded object shapes, \textit{allowing
inpainting also on the level of masks.} With just a single expert, our model can thus
be understood as a generative model extension of \monet{}.

\subsection{Competition mechanism}
\label{sec:competition_mechanism}
For $K>1$, i.e., when explicitly modelling object classes with separate experts,
the objective \eqref{eq:learning-objective} cannot be optimised directly because
it is conditioned on the object identities $k_{1:T}$. To address this issue, we
use the following competition mechanism between experts.

At each inference step $t = T, \ldots, 1$, we apply all experts $(k_t = 1, \ldots, K)$ to the current input
$(\x,\s_t)$ and declare that expert the winner which yields the best competition objective (see below).\footnote{Applying all $K$ experts can be easily parallelised.}
We then use the winning expert $\Hat{k}_t$ to reconstruct the selected scene
component using
\begin{equation*}
  \x_t = \mu_{\theta_{\Hat{k}_t}}\big(\tilde{\*z}_t(\Hat{k}_t)\big),
\end{equation*}
where $\tilde{\z}_t(\Hat{k}_t)$ is encoded from the region
$\tilde{\rr}_{\psi_{\Hat{k}_t}}$ attended by the winning expert $\Hat{k}_t$. We
then compute the contribution to \eqref{eq:learning-objective} from step $t$
assuming fixed $\Hat{k}_t$, and use it to update the winning expert with a
gradient step.
Finally, we update the scope using the winning expert,
\begin{equation}
  \s_{t-1} = \s_t \odot \big( \*1 - \tilde{\rr}_{\psi_{\Hat{k}_t}}(\x, \s_t) \big),
  \label{eq:update_scope}
\end{equation}
to allow for inpainting within the explained region in the following
inference (decomposition) steps.\footnote{To ensure that the entire scene is
explained in $T$ steps, we use the final scope $\*s_1$ as attention region for
all experts in the last inference step $(t=1)$, as also done in \genesis\ and
\monet.}

This competition process can be seen as a \emph{greedy approximation to
maximising} \eqref{eq:learning-objective} w.r.t. $k_{1:T}$. While considering
all possible object combinations would require $O(K^T)$ steps, our competition
procedure is linear in the number of object classes and runs in $O(K\cdot T)$
steps. By choosing an expert at each step $t=T, \ldots, 1$, we approximate the
expectation w.r.t. $q_{\psi}(\s_{t} \given \x, k_{(t+1):T})$---which entangles
the different composition steps and makes inference intractable---using $\s_T = \*1$ and the updates in \eqref{eq:update_scope}.

\paragraph{Competition objective}
While model parameters are updated using the learning objective
\eqref{eq:learning-objective} derived from the ELBO, the choice of competition
objective is ours. Since we use competition to drive specialisation of experts
on different object classes and to greedily infer $k_t$, (i.e., the identity of
the current foreground object), the competition objective should 
reflect such differences between object classes. Object classes can differ
in many ways (shape, color, size, etc) and to different extents, so  the choice of
competition objective is \emph{data-dependent} and may be informed by
prior knowledge.

For instance, in the setting depicted in Figure \ref{fig:samples} where both
color and shape are class-specific, we found that using a combination of
$\Hat{\LL}_{\x,t}$ and $\Hat{\LL}_{\rr_t}$ worked well. However, on the same
data with randomised color (as used in the experiments in
\cref{sec:experiments_and_results}) it did not: due to the greedy optimisation
procedure, the expert which is initially best at reconstructing a particular
color continues to win the competition for explaining regions of that color and
thus receives gradient updates to reinforce this specialisation; such undesired
specialisation corresponds to a local minimum in the optimisation landscape and
can be very hard for the model to escape.

We thus found that relying solely on  $\Hat{\LL}_{\rr, t}$ as the competition
objective (i.e., the reconstruction of the attention region) helps to direct
specialisation towards objects categories. In this case, experts are chosen
based on how well they can model \emph{shape}, and only those experts which can
easily reconstruct (the shape of) a selected region within the current scope
will do well at any given step, meaning that the selected region corresponds to
a foreground object.

Moreover, we found that using a stochastic, rather than deterministic form of
competition, (i.e., experts win the competition with the probabilities
proportional to their competition objectives at a given step) helped
specialisation. In particular, such approach helps prevent the collapse of the
experts in the initial stages of training.

Formally, the probability of expert $k$ winning the
competition is
\begin{equation*}
  P(\Hat{k}_t = k) \propto \exp \left\{-\big( \lambda \Hat{\LL}_{\x,t}(k) + \Hat{\LL}_{\rr,t}(k)\big) \right\},
\end{equation*}
with $\Hat{\LL}_{\x,t}(k)$ and $\Hat{\LL}_{\rr,t}(k)$ being the terms in
\eqref{eq:learning-objective} at step $t$ for an expert $k$. The hyper-parameter
$\lambda$ controls the relative influence of the appearance and shape
reconstruction objectives to make the \emph{data-dependent}
assumptions about the competition mechanism as discussed above.

\section{Experimental results}
\label{sec:experiments_and_results}

To explore \econ{}'s ability to decompose and generate new scenes, we conduct
experiments on synthetic data consisting of colored 2D objects or sprites
(triangles, squares and circles) in different occlusion arrangements. We refer
to Appendix \ref{app:experimental_details} for a detailed account of the used data set, model
architecture, choice of hyperparameters, and experimental setting.
Further experiments can be found in Appendix \ref{app:additional_results}.

\begin{figure}[]
\centering
\centering
\includegraphics[width=.55\textwidth, trim={0 0 0 .75em}, clip ]{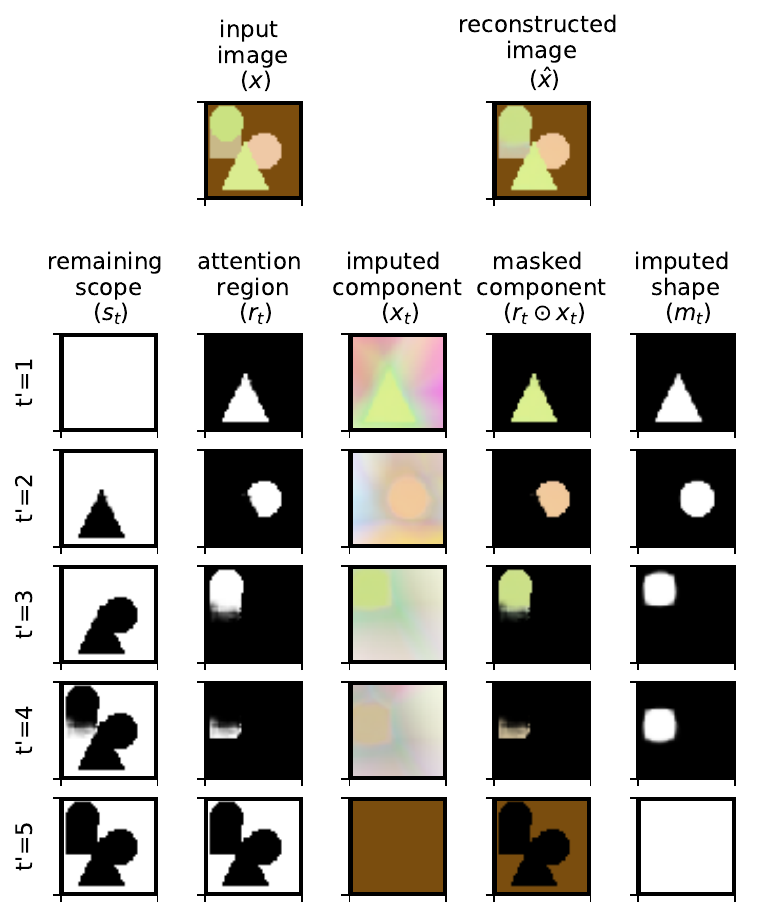}
 \caption{By explaining away a scene \emph{from front to back},  \econ{}  can impute  occluded components $\x_t$ (third column) and---crucially for layered generation and recombination---their shapes $\m_t$ (fifth column) within the already explained regions $\s_t$ (first column). Each inference step $(t'=T+1-t)$ shows only the winning expert's output.}
\label{fig:4_objects_tricky_color}
\end{figure}

\paragraph{\econ{} decomposes scenes and inpaints occluded objects}
Fig.~\ref{fig:4_objects_tricky_color} shows an example of how \econ{} decomposes a
scene with four objects. 
At each inference step, the winning expert segments a region (second col.) within the unexplained part of the image (first col.), and reconstructs the selected object within the attended region (fourth col.).
A distinctive feature of our model is that, despite occlusion, the full shape (rightmost col.) of every object is imputed  (e.g., at step $t'=4$).
This ability to infer complete shapes is a consequence of the assumed layer-wise generative model which manifests itself in our objective via the unconstrained shape reconstruction term \eqref{eq:mask_KL}.

\paragraph{\econ{} generalizes to novel scenes \,}
Fig.~\ref{fig:4_objects_tricky_color}  also illustrates that that the model is capable of decomposing scenes containing multiple objects of the same category, as well as multiple objects of the same color in separate steps.
It does so for a scene with four objects, despite being trained on scenes containing only three objects, one from each class.

\begin{figure}
\centering
\centering
\includegraphics[width=.75\textwidth]{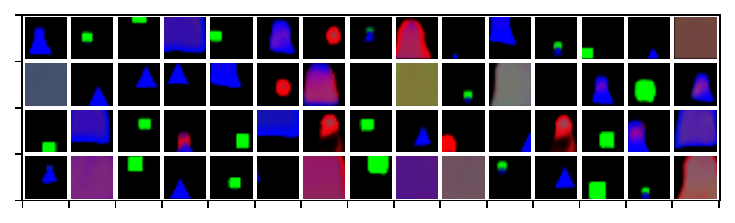}
\caption{Random samples from a single expert (akin to a generative extension of \monet) trained on the data from Fig.~\ref{fig:samples} with ground-truth masks provided. The model learns to separately generate unoccluded objects and background, but lacks control over which object class is sampled.}
\label{fig:single_expert}
\end{figure}

\paragraph{Single expert as generative extension of \monet}
We also investigate training a single expert which we claim to be akin to a generative extension of \monet. When trained on the data from Fig.~\ref{fig:samples} with ground truth masks provided, the expert learns to inpaint occluded shapes and objects as can be seen from the samples in Fig.~\ref{fig:single_expert}. However, all object classes are represented in a shared latent space so that different classes cannot be sampled controllably. 

\begin{figure}
\centering
\includegraphics[width=.75\textwidth]{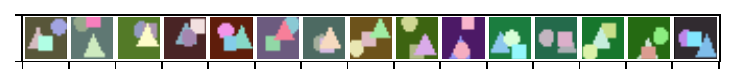}\\
\vspace{-.5em}
\includegraphics[width=.75\textwidth]{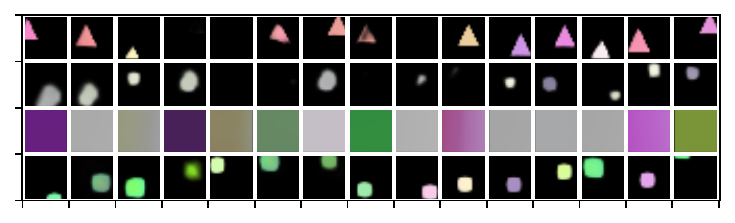}
\caption{Samples from individual experts trained on toy data with random colors (shown in top panel).
Experts (corresponding to rows in the bottom panel) specialise on triangles, circles, background, and squares, respectively, but such specialisation based-purely on shape is significantly harder when color is lost as a powerful cue. This is reflected, e.g., in the imperfect separation between squares and circles, cf. Fig.~\ref{fig:samples}. }
\label{fig:expert_samples_masked}
\end{figure}

\paragraph{Multiple experts specialise on different object classes}
Fig.~\ref{fig:samples}B shows samples from each of the four experts
trained on a dataset with uniquely colored objects (Fig.~\ref{fig:samples}A).
The samples from each expert contain either the same object in different spatial
positions or differently coloured background, indicating that the experts specialised on the
different object classes composing these scenes.

Fig.~\ref{fig:expert_samples_masked} shows the same plot for a model trained
on  scenes consisting of randomly colored objects.
This setting is considerably more challenging because experts have to specialise purely based on shape while also representing color variations.
Yet, experts specialise on different object classes: samples in Fig.~\ref{fig:expert_samples_masked} are
either randomly colored background or objects from mostly one class with different colours and
spatial positions, indicating that the \econ{} is capable of
representing the scenes as compositions of distinct objects in an unsupervised
way.

\begin{figure}
\centering
\includegraphics[width=0.65\textwidth]{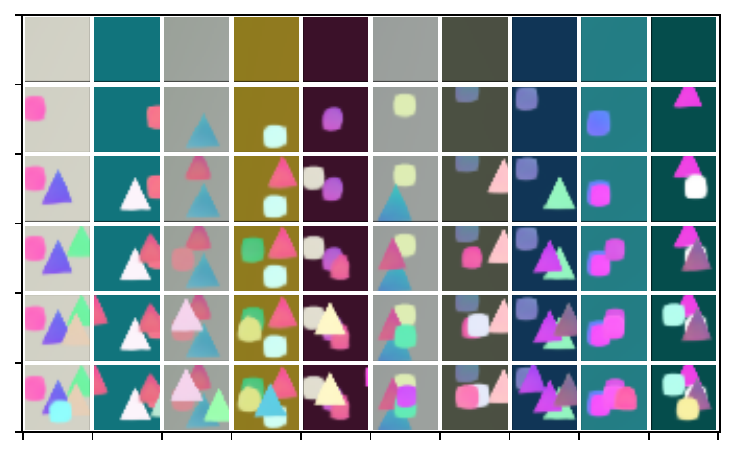}
\caption{Illustration of layer-wise sampling from \econ{} after training on our toy data with random colors.
Starting with a background sample, subsequent rows correspond to sampling additional objects by randomly choosing one of the specialised object experts.}
\label{fig:sample_evolution}
\end{figure}

\paragraph{Controlled and layered generation of new scenes}

The specialisation of experts allows us to controllably generate new scenes with
specific properties. To do so, we follow the sequential generation procedure described in
\cref{sec:layer_based_model} by sampling from one of the experts at each time
step. The number of generation steps $T$, as well as the choices of experts $k_{1:T}$ allow to control the total number and  categories of objects in the generated scene.

Fig.~\ref{fig:samples}C shows samples generated using the experts in
Fig.~\ref{fig:samples}B. In Fig.~\ref{fig:sample_evolution} we show another
example where more and more randomly colored objects are sequentially added.
Even though the generated scenes are quite simple, we believe this result is important as
the ability to generate scenes in a controlled way is a distinctive feature of
our model, which current generative scene models lack.

\section{Discussion}
\label{sec:discussion}

\paragraph{Model assumptions}
While \econ{} aims at modelling scene composition in a faithful way, we make
a number of assumptions for the sake of tractable inference, which need to be revisited
when moving to more general environments. We assume a known (maximum) number of
object classes $K$ which may be restrictive for realistic settings, and choosing $K$ too small may force each expert to represent multiple object classes.
Other assumptions are that the pixel values are modelled as normally distributed, even
though they are discrete in the range $\{0,\ldots,255\}$, and that pixels are conditionally independent given shapes and objects.

\paragraph{Shared vs. object-specific representations}
Recent work on unsupervised representation learning
\citep{bengio2013representation} has largely focused on disentangling factors of
variation within a single shared representation space, e.g., by training a large
encoder-decoder architecture with different forms of regularization
\citep{higgins2017beta, kim2018disentangling, chen2018isolating,
locatello2019challenging}. This is motivated by the observation that certain
(continuous) attributes such as position, size, orientation or color are general
concepts which transcend object-class boundaries. However, the range of values
of these attributes, as well as other (discrete) properties such as shape, can
strongly depend on object class. In this work, we investigate the other extreme
of this spectrum by learning entirely object-specific representations. Exploring
the more plausible middle ground combining both shared and object-specific
representations is an attractive direction for further research.

\paragraph{Extensions and future work}
The goal of decomposing visual scenes into their constituents in an
unsupervised manner from images alone will likely remain a long standing goal of visual
representation learning. We have presented a model that recombines earlier ideas
on layered scene compositions, with more recent models of larger
representational power, and unsupervised attention models. The focus of this work
is to establish physically plausible compositional models for an easy class of
images and to propose a model that naturally captures object-specific
specialization.

With \econ{} and other models as starting point, a number of extensions are
possible. One direction of future work deals with incorporating additional
information about scenes. Here, we consider static, semantically-free
images. Optical flow and depth information can be cues to an attention
process, facilitating segmentation and specialization. First results in the
direction of video data have been shown by~\citet{xu2019unsupervised}.  Natural
images typically carry semantic meaning and objects are not ordered in arbitrary
configurations. Capturing dependencies between objects (e.g., using an auto-regressive prior over depth ordering as in \genesis),
albeit challenging, could help disambiguate between scene
components. Another direction of future work is to relax the unsupervised
assumption, e.g., by exploring a semi-supervised approach, which might help
improve stability.

On the modelling side, extensions to recurrent architectures and iterative
refinement as in \iodine{} appear promising. Our model entirely separates experts from each other but, depending on object similarity, one can also
include shared representations which will help transfer already learned
knowledge to new experts in a continual learning scenario.

\section{Conclusion}
\label{sec:conclusion}
While the scenes studied here and in the recent works of~\citet{MONET, IODINE,
GENESIS} are still in stark contrast to the impressive results that holistic generative
models are able to achieve, we believe it is
the right time to revisit the unsupervised scene composition problem.
Our goal is to build re-combineable systems, where different components
can be used for new scene inference tasks. In the spirit of the
analysis-by-synthesis approach, this requires the ability to re-create
physically plausible visual scenes. Disentangling the scene formation process
from the objects is one crucial component thereof, and the vast number of object
types will require the ability of unsupervised learning from visual input alone.

\section*{Acknowledgements}
The authors would like to thank Alex Smola, Anirudh Goyal, Muhammad Waleed Gondal, Chris Russel, Adrian Weller, Neil Lawrence, and the Empirical Inference ``deep learning \& causality'' team at the MPI for Intelligent Systems for helpful discussions and feedback.

M.B. and B.S. acknowledge support from the German Science Foundation (DFG) through the CRC 1233 ``Robust Vision'' project number 276693517, the German Federal Ministry of Education and Research (BMBF) through the Tübingen AI Center (FKZ: 01IS18039A), and the DFG Cluster of Excellence ``Machine Learning – New Perspectives for Science'' EXC 2064/1, project number 390727645.

\clearpage
\bibliography{scenes_iclr}
\bibliographystyle{iclr2020_conference}

\clearpage
\appendix
\numberwithin{equation}{section}
\section{Derivations}
\label{app:derivations}
\subsection{Derivation of ELBO}

We now provide a detailed derivation of the evidence lower bound (ELBO) used in the main paper.
For ease of notation we use vector notation and omit explicitly summing over pixel- and latent dimensions (as done in the implementation).

We start by writing $p_\theta(\*x|k_{1:T})$ as an expectation w.r.t. $q$ using importance sampling as follows:
\begin{align*}
p_\theta(\*x|k_{1:T})
&=
\mathbb{E}_{p_\theta(\*r_{1:T}, \*z_{1:T}| k_{1:T})}
\Big[
p_\theta(\*x| \*r_{1:T}, \*z_{1:T}, k_{1:T})
\Big]\\
 &=
 \mathbb{E}_{q_{\phi,\psi}(\*r_{1:T}, \*z_{1:T} |\*x, k_{1:T})}
 \bigg[
 \frac{p_\theta(\*x, \*r_{1:T}, \*z_{1:T}| k_{1:T})}{q_{\phi,\psi}(\*r_{1:T}, \*z_{1:T} |\*x, k_{1:T})}
 \bigg].
\end{align*}

Applying the concave function $\log(\argdot)$ and using Jensen's inequality we obtain
\begin{equation}
\label{eq:ELBO_Jensen}
\log p_\theta(\*x|k_{1:T})
\geq
\mathbb{E}_{q_{\phi,\psi}(\*r_{1:T}, \*z_{1:T} |\*x, k_{1:T})}
\bigg[
\log \frac{p_\theta(\*x, \*r_{1:T}, \*z_{1:T}| k_{1:T})}{q_{\phi,\psi}(\*r_{1:T}, \*z_{1:T} |\*x, k_{1:T})}
\bigg].
\end{equation}

Using the chain rule of probability and properties of $\log(\argdot)$, we can rearrange the integrand on the RHS of \eqref{eq:ELBO_Jensen} as
\begin{equation}\\
 \label{eq:ELBO_terms}
 \log p_\theta(\*x| \*r_{1:T}, \*z_{1:T}, k_{1:T})
- \log \frac{q_{\psi}(\*r_{1:T}|\*x, k_{1:T})}{p_\theta(\*r_{1:T}| \*z_{1:T}, k_{1:T})}
- \log \frac{q_{\phi}(\*z_{1:T} |\*x, \*r_{1:T}, k_{1:T})}{p_\theta(\*z_{1:T}| k_{1:T})}.
\end{equation}

We will consider the three terms in \eqref{eq:ELBO_terms} separately and define their expectations w.r.t. the approximate posterior as
\begin{align*}
\mathcal{L}_{\*x}(\theta, \psi, \phi|k_{1:T})&:= \mathbb{E}_{q_{\phi,\psi}(\*r_{1:T}, \*z_{1:T} |\*x, k_{1:T})}
 \bigg[
  \log p_\theta(\*x| \*r_{1:T}, \*z_{1:T}, k_{1:T})
 \bigg],\\
 \mathcal{L}_{\*r}(\theta, \psi, \phi|k_{1:T})&:= \mathbb{E}_{q_{\phi,\psi}(\*r_{1:T}, \*z_{1:T} |\*x, k_{1:T})}
 \bigg[
 - \log \frac{q_{\psi}(\*r_{1:T}|\*x, k_{1:T})}{p_\theta(\*r_{1:T}| \*z_{1:T}, k_{1:T})}
 \bigg],\\
 \mathcal{L}_{\*z}(\theta, \psi, \phi|k_{1:T})&:= \mathbb{E}_{q_{\phi,\psi}(\*r_{1:T}, \*z_{1:T} |\*x, k_{1:T})}
 \bigg[
 - \log \frac{q_{\phi}(\*z_{1:T} |\*x, \*r_{1:T}, k_{1:T})}{p_\theta(\*z_{1:T}| k_{1:T})}
 \bigg].
\end{align*}

Next, we use our modelling assumptions stated in the paper to simplify these terms, starting with $\mathcal{L}_{\*z}$.

Using the assumed factorisation of the approximate posterior, in particular $q_{\psi}(\*r_{t}|\*x, \*r_{(t+1):T}, k_{t})=q_{\psi}(\*r_{t}|\*x, \*s_t, k_{t})$, as well as the fact that $p_\theta(\*z_t|k_t)=p(\*z)$, splitting the expectation into two parts, and using linearity of the expectation operator, we find that $\mathcal{L}_{\*z}$ can be written as follows:
\begin{align*}
 \mathcal{L}_{\*z}(\psi, \phi|k_{1:T})
 &=
 \mathbb{E}_{q_{\psi}(\*r_{1:T}|\*x, k_{1:T})}
    \Bigg[
 \mathbb{E}_{q_{\phi}(\*z_{1:T}|\*x, \*r_{1:T}, k_{1:T})}
 \bigg[
 - \sum_{t=1}^T
  \log \frac{q_{\phi}(\*z_{t} |\*x, \*r_{t}, k_{t})}{p(\*z)}
 \bigg]
   \Bigg]\\
 &=
 -\sum_{t=1}^T
 \mathbb{E}_{q_{\psi}(\*r_{t:T}|\*x, k_{t:T})}
  \Bigg[
  \mathbb{E}_{q_{\phi}(\*z_{t}|\*x, \*r_{t}, k_{t})}
   \bigg[
 \log \frac{q_{\phi}(\*z_{t} |\*x, \*r_{t}, k_{t})}{p(\*z)}
  \bigg]
   \Bigg]\\
   &=
    -\sum_{t=1}^T
    \mathbb{E}_{q_{\psi}(\*s_{t}|\*x, k_{(t+1):T})}
 \mathbb{E}_{q_{\psi}(\*r_{t}|\*x, \*s_t, k_{t})}
  \Bigg[
\KL[\Big]{q_{\phi}(\*z_{t}|\*x, \*r_{t}, k_{t})}{ p(\*z)}
   \Bigg].
\end{align*}

Next, we consider $\mathcal{L}_{\*r}$. Using a similar argument as for $\mathcal{L}_{\*z}$, we find that
\begin{align*}
\mathcal{L}_{\*r}(\theta, \psi, \phi|k_{1:T})
&=
 \mathbb{E}_{q_{\psi}(\*r_{1:T} |\*x, k_{1:T})}
 \Bigg[
 \mathbb{E}_{q_{\phi}(\*z_{1:T} |\*x, \*r_{1:T}, k_{1:T})}
 \bigg[
 - \sum_{t=1}^T
 \log \frac{q_{\psi}(\*r_{t}|\*x,\*s_t, k_{t})}{p_\theta(\*r_{t}|\*s_t, \*z_{t}, k_{t})}
 \bigg]
 \Bigg]\\
 &=
 - \sum_{t=1}^T
 \mathbb{E}_{q_{\psi}(\*s_{t}|\*x, k_{(t+1):T})}
 \mathbb{E}_{q_{\psi}(\*r_{t} |\*x,\*s_t, k_{t})}
 \Bigg[
 \mathbb{E}_{q_{\phi}(\*z_{t} |\*x, \*r_{t}, k_{t})}
 \bigg[
 \log \frac{q_{\psi}(\*r_{t}|\*x,\*s_t, k_{t})}{p_\theta(\*r_{t}|\*s_t, \*z_{t}, k_{t})}
 \bigg]
 \Bigg].
\end{align*}

Finally, we consider $\mathcal{L}_{\*x}$.
Substituting the  Gaussian likelihood for $p_\theta(\*x| \*r_{1:T}, \*z_{1:T}, k_{1:T})$, ignoring constants which do not depend on any learnable parameters, and using the fact that $\*r_t$ is binary and $\sum_{t=1}^T\*r_t=\*1$, we obtain
\begin{align*}
\mathcal{L}_{\*x}(\theta, \psi, \phi|k_{1:T})
&=
\mathbb{E}_{q_{\psi}(\*r_{1:T} |\*x, k_{1:T})}
 \Bigg[
 \mathbb{E}_{q_{\phi}(\*z_{1:T} |\*x, \*r_{1:T}, k_{1:T})}
 \bigg[
-\frac{1}{2\sigma_{\*x}^2}
\norm[\Big]{\*x - \sum_{t=1}^T \*r_t \odot\mu_{\theta_{k_t}}(\*z_t)}^2
 \bigg]
 \Bigg]\\
 &=
 -\frac{1}{2\sigma_{\*x}^2}
 \mathbb{E}_{q_{\psi}(\*r_{1:T} |\*x, k_{1:T})}
 \Bigg[
 \mathbb{E}_{q_{\phi}(\*z_{1:T} |\*x, \*r_{1:T}, k_{1:T})}
 \bigg[
\sum_{t=1}^T \*r_t \odot
\norm[\big]{
\*x - \mu_{\theta_{k_t}}(\*z_t)}^2
 \bigg]
 \Bigg]\\
 &=
 - \sum_{t=1}^T
 \frac{1}{2\sigma_{\*x}^2}
 \mathbb{E}_{q_{\psi}(\*s_{t}|\*x, k_{(t+1):T})}
  \mathbb{E}_{q_{\psi}(\*r_{t} |\*x,\*s_t, k_{t})}
 \Bigg[
 \*r_t \odot
 \mathbb{E}_{q_{\phi}(\*z_{t} |\*x, \*r_{t}, k_{t})}
 \bigg[
\norm[\big]{\*x - \mu_{\theta_{k_t}}(\*z_t)}^2
 \bigg]
 \Bigg],
\end{align*}
where $\norm{\argdot}^2$ denotes the pixel-wise L2-norm between two RGB vectors. (Recall that $\mathcal{L}_{\*x}$, $\mathcal{L}_{\*r}$, and $\mathcal{L}_{\*z}$ are defined as quantities in $\mathbb{R}^{D}$, $\mathbb{R}^{D}$, and $\mathbb{R}^{L}$, respectively, and that summation over these dimensions yields the desired scalar objective.)

We observe that $\mathcal{L}_{\*x}$, $\mathcal{L}_{\*r}$, and $\mathcal{L}_{\*z}$ can all be written as sums over the $T$ composition steps.

We thus define:
\begin{align*}
\mathcal{L}_{\*x, t}(\theta, \psi, \phi|k_t)
&:=
 \frac{1}{2\sigma_{\*x}^2}
  \mathbb{E}_{q_{\psi}(\*r_{t} |\*x,\*s_t, k_{t})}
 \Bigg[
 \*r_t \odot
 \mathbb{E}_{q_{\phi}(\*z_{t} |\*x, \*r_{t}, k_{t})}
 \bigg[
\norm[\big]{\*x - \mu_{\theta_{k_t}}(\*z_t)}^2
 \bigg]
 \Bigg],
\\
\mathcal{L}_{\*r, t}(\theta, \psi, \phi|k_t)
&:=
\mathbb{E}_{q_{\psi}(\*r_{t} |\*x,\*s_t, k_{t})}
 \Bigg[
 \mathbb{E}_{q_{\phi}(\*z_{t} |\*x, \*r_{t}, k_{t})}
 \bigg[
 \log \frac{q_{\psi}(\*r_{t}|\*x,\*s_t, k_{t})}{p_\theta(\*r_{t}|\*s_t, \*z_{t}, k_{t})}
 \bigg]
 \Bigg],
 \\
 \mathcal{L}_{\*z, t}(\psi, \phi|k_t)
&:=
\mathbb{E}_{q_{\psi}(\*r_{t}|\*x, \*s_t, k_{t})}
  \Bigg[
\KL[\Big]{q_{\phi}(\*z_{t}|\*x, \*r_{t}, k_{t})}{ p(\*z)}
   \Bigg],\\
   \mathcal{L}_t(\theta, \psi, \phi|k_t)
   &:=
   -  \mathbb{E}_{q_{\psi}(\*s_{t}|\*x, k_{(t+1):T})}
 \Big( \mathcal{L}_{\*x,t}(\theta, \psi, \phi|k_t) + \mathcal{L}_{\*z,t}(\psi, \phi|k_t) + \mathcal{L}_{\*r,t}(\theta, \psi, \phi|k_t)\Big)
\end{align*}

Finally, it then follows that
\begin{equation*}
\mathcal{L}(\theta, \psi, \phi|k_{1:T})
:=
\sum_{t=1}^T \mathcal{L}_{t}(\theta, \psi, \phi|k_t)
\leq
\log p_\theta(\*x|k_{1:T})
\end{equation*}

\subsection{Derivation of generative region distribution}
We now derive the distribution in \eqref{eq:generative_region_dist}. We will use the fact that $\rr_t=\m_t \odot \s_t$, and that$\s_t$ can be written as $\s_t=\mathbf{1} - \sum_{t'=t+1}^T \rr_t'$, as well as the conditional independencies implied by our model, see Figure \ref{fig:graphical_model}.
Considering the pixel-wise distribution and marginalising over $\m_t$, we obtain:
\begin{align*}
p_\theta\left(\rr_t=1|k_t,\z_t, \rr_{(t+1):T}\right)
&=
\sum_{\m_t=0}^1 p_\theta\left(\rr_t=1|\m_t, k_t,\z_t, \rr_{(t+1):T}\right) p_\theta\left(\m_t|k_t,\z_t, \rr_{(t+1):T} \right) \\
&=
\sum_{\m_t=0}^1 p_\theta\left(\rr_t=1|\m_t, \s_t\right) p_\theta\left(\m_t|k_t,\z_t \right) \\
&=
0 + p_\theta\left(\rr_t=1|\m_t=1, \s_t \right) p_\theta\left(\m_t=1|k_t,\z_t\right)\\
&=\s_t\odot \tilde{\m}_{\theta_{k_t}}(\z_t).
\end{align*}

Since $\rr_t$ is binary, this fully determines its distribution.

\clearpage
\section{Additional experimental results}
\label{app:additional_results}
\begin{figure}[htb!]
\centering
\begin{subfigure}{0.45\textwidth}
\centering
\includegraphics[width=\textwidth]{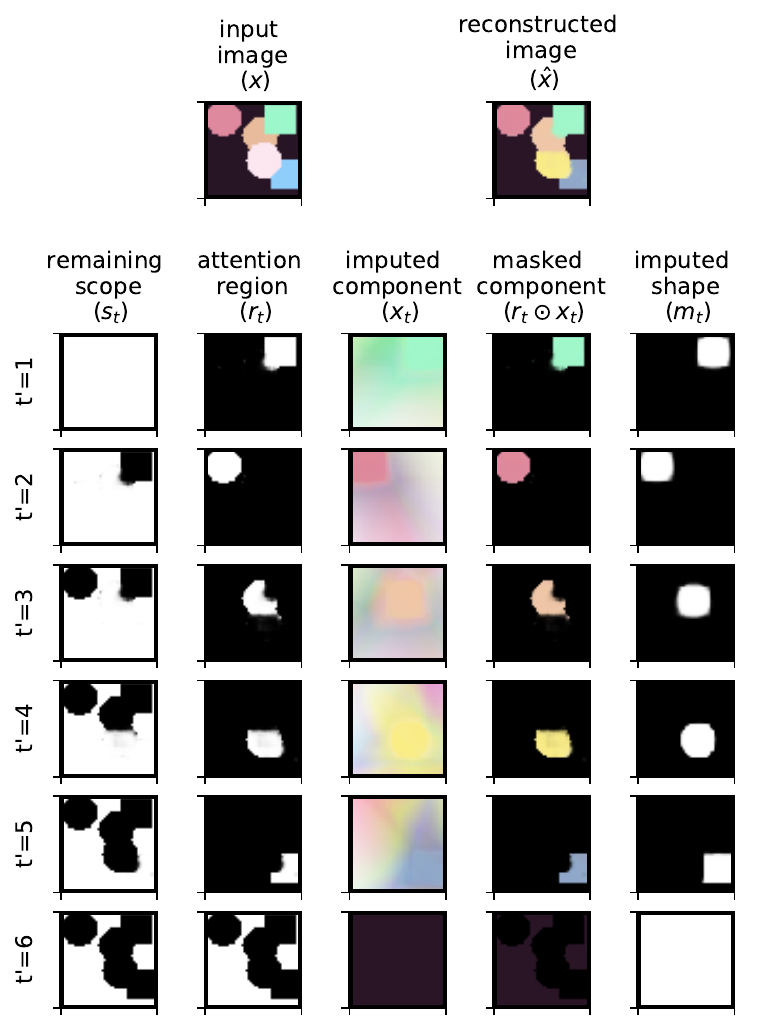}
\caption{}
\end{subfigure}
\hfill
\begin{subfigure}{0.45\textwidth}
\centering
\includegraphics[width=\textwidth]{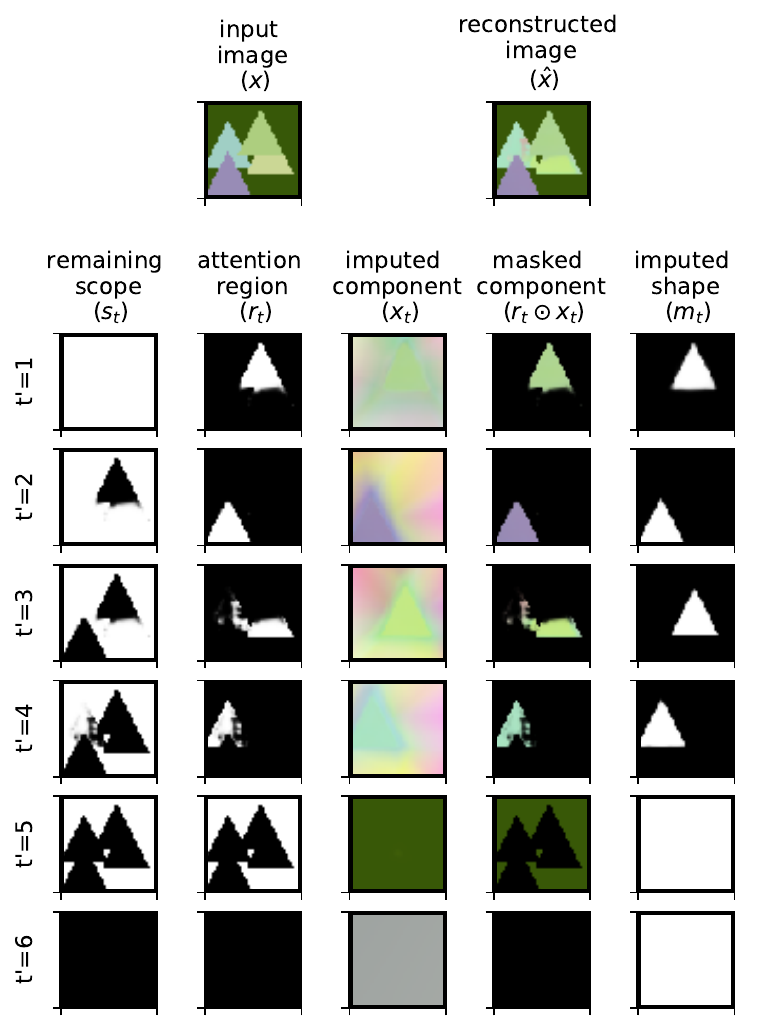}
\caption{}
\end{subfigure}

\vspace{2em}
\begin{subfigure}{0.45\textwidth}
\centering
\includegraphics[width=\textwidth]{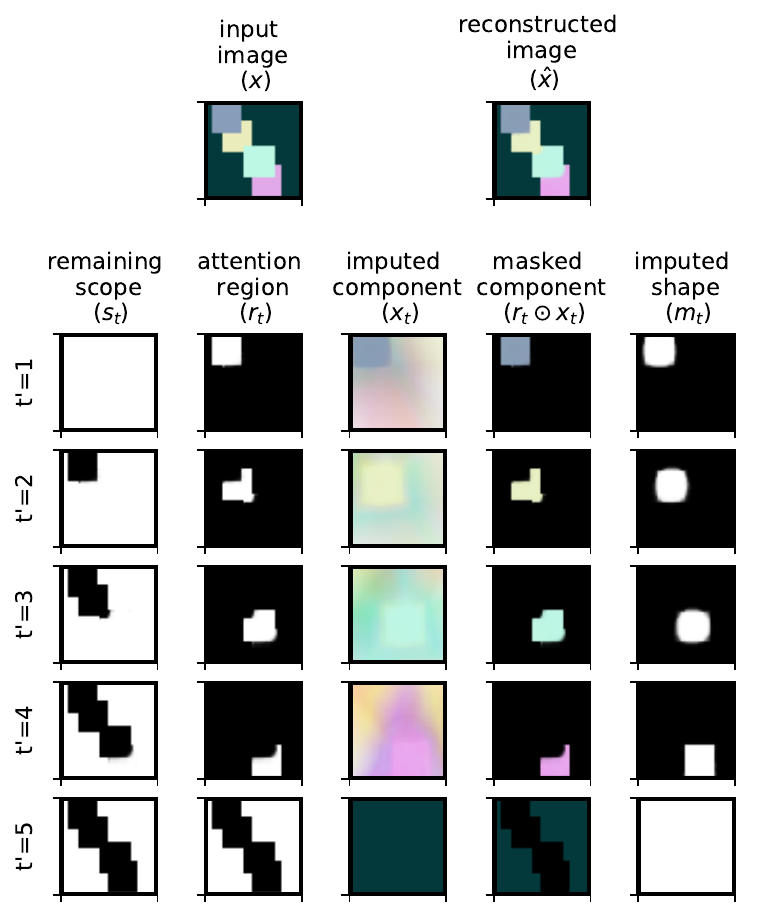}
\caption{}
\end{subfigure}
\hfill
\begin{subfigure}{0.45\textwidth}
\centering
\includegraphics[width=\textwidth]{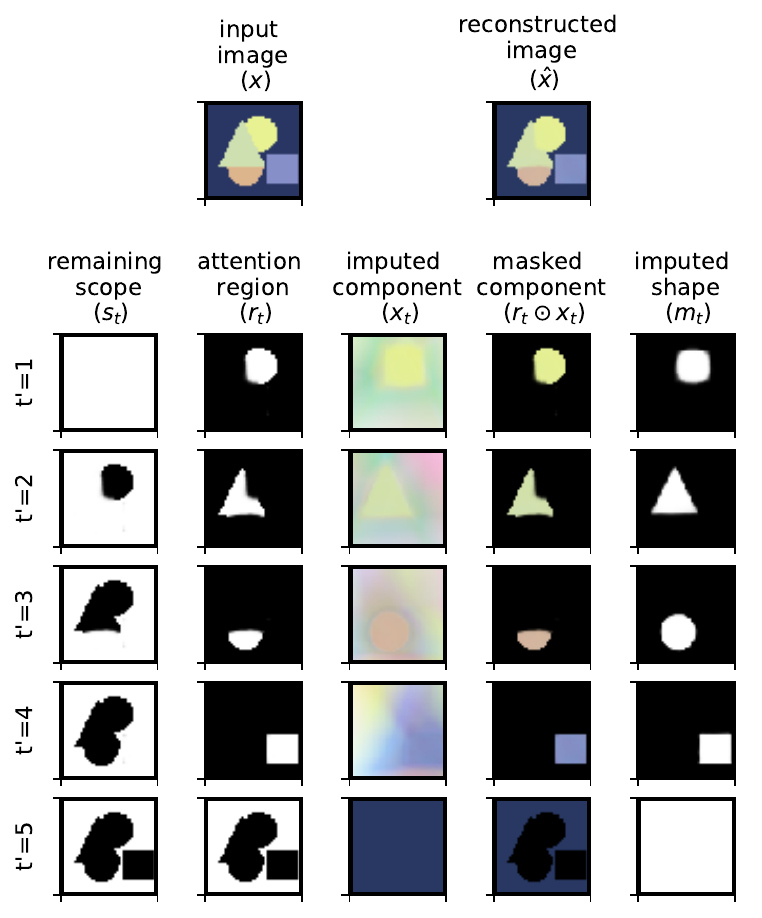}
\caption{}
\end{subfigure}
\caption{Additional decomposition plots for o.o.d.\ data. The model was trained with four experts on scenes containing three objcets (one triangle, square, and circle each) arranged in random order. }
\label{fig:app_decomposition}
\end{figure}

Figure \ref{fig:app_decomposition} shows four additional examples of \econ{}  decomposing scenes consisting of multiple randomly coloured shapes. The model was trained on the data from Fig.~\ref{fig:expert_samples_masked}, but is able to decompose scenes with five objects (a), multiple occluding objects from the same class (b, c), and objects of similar color to the background (d). Moreover, (b) suggests that additional timesteps ($t'=6$) are simply ignored if they are not needed.

\clearpage
\section{Experimental details}
\label{app:experimental_details}
\subsection{Datasets}

\paragraph{Synthetic dataset: uniquely colored objects} The dataset consists of
images of circles, squares and triangles on a randomly and uniformly colored
background, such that there is a unique correspondence between object color and
class identites (red circles, green squares, blue triangles). The background
color is randomly chosen to be an RGB value with each channel being a random
integer between 0 and 127, while the RGB values of the object colors are
(255,0,0), (0,255,0), (0,0,255) for circles, squares and triangles respectively.
The spatial positions of the objects are randomly chosen such that each of the
objects entirely fits into an image without crossing the image boundary.

The models shown in Fig.~1 and 5 have been trained on a version of such dataset
containing images with exactly three objects per image (one of each class) in
random depth orders (Fig.~1, top row). The training and validation splits
include \num{50000} and 100 such images respectively.

\paragraph{Synthetic dataset: randomly colored objects} This dataset is the same
as the one described above with the difference that the objects (circles,
squares and triangles) are randomly colored with the corresponding RGB values
being random integers between 128 and 255.

The models shown in Fig.~4, 6 and 8 have been trained on a version of such
dataset containing images with exactly three objects per image (one of each
class) in random depth orders (Fig.~6, top row). The training and validation
splits include 50000 and 100 such images respectively.

\subsection{Architecture details}

Each expert in our model consists of attention network computing the
segmentation regions as a function of the input image and the scope at a given
time step, and a VAE reconstructing the image appearance within the segmentation
region and inpainting the unoccluded shape of object. Below we describe the
details of architectures we used for each of the expert networks.

\subsubsection{Expert VAEs}

\paragraph{Encoder} The VAE encoder consists of multiple blocks, each of which
is composed of $3 \times 3$ convolutional layer, ReLU non-linearity, and $2
\times 2$ max pooling. The output of the final block is flattened and
transformed into a latent space vector by means of two fully connected layers.
The output of the first fully-connected layer has 4 times the number of latent
dimensions activations, which are passed through the ReLU activation, and
finally linearly mapped to the latent vector by a second fully-connected layer.

\paragraph{Decoder} Following \citet{MONET}, we use spatial a broadcast decoder.
First, the latent vector is repeated on a spatial grid of the size of an input
image, resulting in a 3D tensor with spatial dimensions being that of an input,
and as many feature maps as there are dimensions in the latent space. Second, we
concatenate the two coordinate grids (for $x-$ and $y-$coordinates) to this
tensor. Next, this tensor is processed by a decoding network consisting of as
many blocks as the encoder, with each block including a $3 \times 3$
convolutional layer and ReLU non-linearity. Finally, we apply a $1 \times 1$
convolutional layer with sigmoid activation to the output of the decoding
network resulting in an output of 4 channel (RGB + shape reconstruction).

\subsubsection{Attention network}

We use the same attention network architecture as in \citet{MONET} and the
implementation provided by \citet{GENESIS}. It consists of U-Net \citep{UNet}
with 4 down and up blocks consisting of a $3 \times 3$ convolutional layer,
instance normalisation, ReLU activation and down- or up-sampling by a factor of
two. The numbers of channels of the block outputs in the down part (the up part
is symmetric) of the network are: 4 - 32 - 64 - 64 - 64.

\subsection{Training details}

We implemented the model in PyTorch \citep{pytorch}. We use the batch size of
32, Adam optimiser \citep{adam}, and initial learning rate of $5 \cdot 10^{-4}$.
We compute the validation loss every 100 iterations, and if the validation loss
doesn't improve for 5 consecutive evaluations, we decrease the learning rate by
a factor of $\sqrt{10}$. We stop the training after 5 learning rate decrease
step.

\subsection{Cross-validation}

\paragraph{Synthetic dataset: uniquely colored objects} The results in Fig.~1
were obtained by cross-validating 512 randomly sampled architectures with the
following ranges of parameters:

\begin{table}[h]
  \centering
  \begin{tabular}{ll}
    \multicolumn{1}{c}{\textbf{Parameter}}              & \multicolumn{1}{c}{\textbf{Range}} \\ \hline
    Latent dimension                                    & 2 to 3   \\
    Number of layers in encoder and decoder             & 2 to 4   \\
    Number of features per layer in encoder and decoder & 4 to 32  \\
    $\beta$ (KL term weight in (12))                    & 0.5 to 2 \\
    $\gamma$ (shape reconstruction weight in (12))      & 0.1 to 10 \\
    Number of experts                                   & 4 (three objects + background) \\
    Number of time steps                                & 4 (three objects + background)
  \end{tabular}
\end{table}

The best performing model in terms of the validation loss (which is shown in
Fig.~1) has the latent dimension of 2, 4 layers in encoder and decoder, 32
features per layer, $\beta = 9.54$ and $\gamma = 0.52$.

The results in Fig.~5 were obtained using the same model as above but with one
expert.

\paragraph{Synthetic dataset: randomly colored objects} The results in Figs.~4,
6, and 7 were obtained by cross-validating 512 randomly sampled architectures
with the following ranges of parameters:

\begin{table}[h]
  \centering
  \begin{tabular}{ll}
    \multicolumn{1}{c}{\textbf{Parameter}}              & \multicolumn{1}{c}{\textbf{Range}} \\ \hline
    Latent dimension                                    & 4 to 5   \\
    Number of layers in encoder and decoder             & 3 to 4   \\
    Number of features per layer in encoder and decoder & 16 to 32  \\
    $\beta$ (KL term weight in (12))                    & 1 \\
    $\gamma$ (shape reconstruction weight in (12))      & 0.5 to 5 \\
    Number of experts                                   & 4 (three objects + background) \\
    Number of time steps                                & 4 (three objects + background)
  \end{tabular}
\end{table}

The best performing model in terms of the validation loss (which is shown in
Fig.~1) has the latent dimension of 5, 3 layers in encoder and decoder, 32
features per layer, $\beta = 1$ and $\gamma = 3.26$.

\end{document}